\documentclass[letterpaper, 10 pt, conference]{ieeeconf}  

\IEEEoverridecommandlockouts                              

\usepackage{graphics} 
\usepackage{epsfig} 
\usepackage{mathptmx} 
\usepackage{times} 
\usepackage{amsmath} 
\usepackage{amssymb}  
\usepackage{multirow}
\usepackage{booktabs}
\usepackage{xcolor}
\usepackage{pifont}
\usepackage{cite} 
\usepackage{threeparttable}

\newcommand{\cavan}[1]{{\color{blue}(cavan: {#1})}} 

\title{\LARGE \bf
DHAGrasp: Synthesizing Affordance-Aware Dual-Hand Grasps \\ with
Text Instructions
}

\author{
Quanzhou Li$^{1}$, Zhonghua Wu$^{2}$, Jingbo Wang$^{3}$, Chen Change Loy$^{1\dag}$, Bo Dai$^{4}$
\thanks{$^{1}$S-Lab, Nanyang Technological University}
\thanks{$^{2}$SenseTime Research}
\thanks{$^{3}$Shanghai AI Laboratory}
\thanks{$^{4}$The University of Hong Kong}
\thanks{$^{\dag}$Corresponding author}
}

\begin{document}

\maketitle
\thispagestyle{empty}
\pagestyle{empty}


\begin{abstract}
Learning to generate dual-hand grasps that respect object semantics is essential for robust hand–object interaction but remains largely underexplored due to dataset scarcity. Existing grasp datasets predominantly focus on single-hand interactions and contain only limited semantic part annotations. To address these challenges, we introduce a pipeline, \textit{SymOpt}, that constructs a large-scale dual-hand grasp dataset by leveraging existing single-hand datasets and exploiting object and hand symmetries. Building on this, we propose a text-guided dual-hand grasp generator, \textit{DHAGrasp}, that synthesizes \textit{D}ual-\textit{H}and \textit{A}ffordance-aware \textit{Grasp}s for unseen objects. Our approach incorporates a novel dual-hand affordance representation and follows a two-stage design, which enables effective learning from a small set of segmented training objects while scaling to a much larger pool of unsegmented data. Extensive experiments demonstrate that our method produces diverse and semantically consistent grasps, outperforming strong baselines in both grasp quality and generalization to unseen objects. The project page is at https://quanzhou-li.github.io/DHAGrasp/.
\end{abstract}

\section{Introduction}
\label{sec:intro}

Defining effective grasp targets is fundamental for enabling robust interaction between hands and objects. While extensive research 
\cite{song2025overview} 
has been conducted on single-hand grasping, more sophisticated tasks often require coordinated dual-hand grasps. Moreover, in practical situations, interaction with specific parts of an object is frequently necessary, calling for a generator that generates coordinated two-hand grasps guided by the functional semantics of object parts rather than geometry along.

Despite the importance of dual-hand approaches, research in this area remains underexplored, primarily due to the scarcity of suitable two-hand grasp datasets. Existing datasets, such as \cite{fan2023arctic} and \cite{Kwon_2021h2o}, include sequences of two-hand manipulations but focus mainly on manipulation motions rather than the grasping phase itself and feature only a limited number of objects with restricted grasp diversity. In the absence of two-hand grasp datasets, approaches such as \cite{shao2024bimanual} employ energy-based optimization to generate dual-hand grasps. These processes, however, tend to be slow and yield relatively low success rates.

\begin{figure}[!t]
  \centering
  \includegraphics[width=\linewidth]{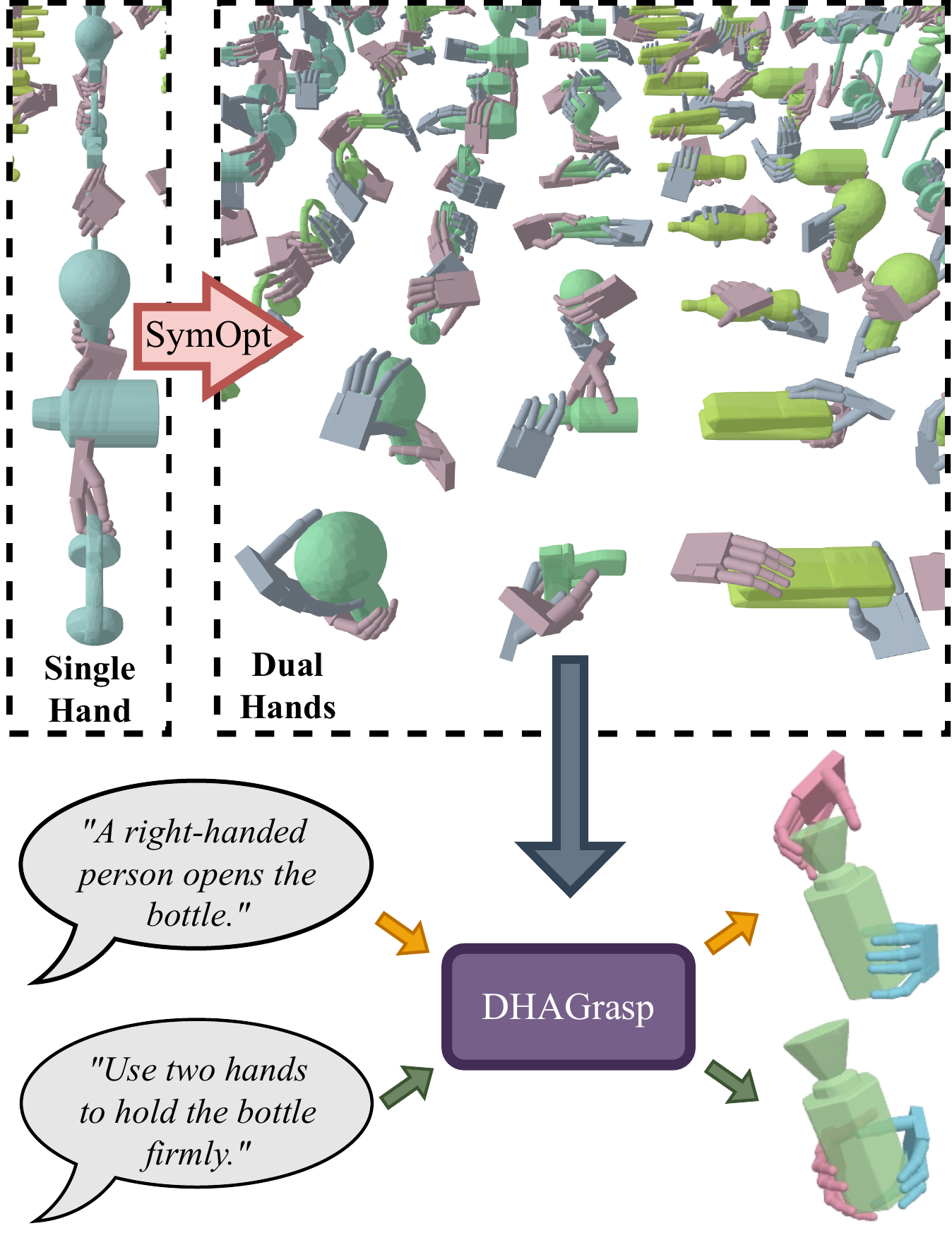}
  \caption{We propose a data generation framework, \textit{SymOpt}, which constructs large-scale two-hand grasp data from existing single-hand grasps. Subsequently, we collect and annotate an affordance-aware dataset and employ it together with our novel generation pipeline, \textit{DHAGrasp}, to synthesize two-hand grasps conditioned on text instructions.
  }
  \label{fig:teaser}
\end{figure}


To overcome these limitations, we propose a novel approach that leverages the abundance of existing single-hand grasp datasets. Specifically, we exploit object and hand symmetries to construct a large-scale dual-hand grasp dataset. Most objects exhibit approximate bilateral symmetry; thus, we mirror ground-truth right-hand grasps across an identified pseudo-symmetric plane to obtain corresponding left-hand grasps. These mirrored grasps are then combined into candidate dual-hand proposals. To ensure physical plausibility, we introduce an optimization scheme, termed \textit{SymOpt}, which eliminates interpenetrations between the hands and the object. Using this pipeline, we derive a large-scale dataset, DualHands-Full, from DexGraspNet \cite{wang2022dexgraspnet}, and further construct DualHands-Sem, a semantics-augmented subset that associates functional object parts with dual-hand grasps. As summarized in Table~\ref{tab:statistics}, our dataset features a substantially larger number of grasp configurations per object compared with prior work and uniquely incorporates semantics into dual-hand grasping.

Although our dataset contains abundant dual-hand grasps, it inherits the limitations of single-hand datasets, namely the limited availability of consistent semantic part labels, which are costly to obtain at scale. Directly training on dual-hand data with such limited segmentation would lead to suboptimal performance. To address this problem, we propose a novel dual-hand contact representation, which bridges the unsegmented objects to the objects with semantics. Our contact representation consists of a contact map, a part map, and most importantly, a set of affordance directions. By leveraging the representation, our approach learns effectively from a small subset of segmented training objects while scaling to a much larger pool of unsegmented data.


With the dual-hand contact representation, we introduce a text-guided Dual-Hand Affordance-aware Grasp (\textit{DHAGrasp}) generator. The key intuition is inspired by how humans recognize the functional parts of novel objects by drawing on prior knowledge of similar items. Specifically, our generator operates in two stages: (1) Text2Dir, a semantics-based affordance module that predicts category-level affordance directions from object geometry and text embeddings; and (2) Dir2Grasp, an affordance-conditioned grasp synthesizer that generates dual-hand grasps aligned with the predicted affordance directions. This design enables effective utilization of limited annotated data while scaling to large unsegmented datasets.

We conduct extensive experiments to evaluate both our datasets and the grasp generation method. Our data generation approach, \textit{SymOpt}, produces datasets an order of magnitude larger than those in previous works and yields substantially higher success rates on the generated grasps. We further demonstrate the effectiveness of our generation pipeline, \textit{DHAGrasp}, which excels in two-hand coordination and outperforms prior methods across multiple metrics.

In summary, our contributions are fourfold. 1) we introduce a pipeline, \textit{SymOpt}, that constructs a dual-hand object grasp dataset leveraging the advances of single-hand datasets. 2) We assemble a large-scale dual-hand grasp dataset DualHands-Full, together with a semantics-based sub-dataset, DualHands-Sem. 3) We propose a novel dual-hand contact representation that enables within-category generalization. 4) We design a Dual-Hand Affordance-aware Grasp Model, \textit{DHAGrasp}, that synthesizes dual-hand grasps based on text instructions.


\definecolor{darkgreen}{RGB}{0,180,0}

\begin{table}[!t]
  \centering
  \caption{Statistics of our datasets compared to previous works.}
  \label{tab:statistics}
  \resizebox{\linewidth}{!}{%
    \begin{tabular}{lcccc}
      \toprule
       & Dual-hand & Num. obj. & Num. grasps & Avg. grasps/obj. \\
      \midrule
      CapGrasp~\cite{li2024semgrasp}  & \textcolor{red}{\ding{55}} & 1,800 & 50,000 & 27 \\
      DexTOG~\cite{zhang2025dextog}   & \textcolor{red}{\ding{55}} & 80   & 80,000  & 1,000 \\
      DexGraspNet~\cite{wang2022dexgraspnet} & \textcolor{red}{\ding{55}} & 5,355 & 1,320,000 & 246 \\
      BimanGrasp~\cite{shao2024bimanual}     & \textcolor{darkgreen}{\checkmark} & 900 & 150,000 & 166 \\
      DualHands-Sem (Ours) & \textcolor{darkgreen}{\checkmark} & 157 & 40,000 & $>$250 \\
      DualHands-Full (Ours) & \textcolor{darkgreen}{\checkmark} & 802 & 1,302,000 & $>$1,600 \\
      \bottomrule
    \end{tabular}
  }
  \vspace{0.4em}
  \begin{minipage}{\linewidth}
    \footnotesize
    \vspace{0.6em}
    \textit{Note}: DualHands-Full is a large, task-agnostic dataset that captures general dual-hand grasp patterns, while DualHands-Sem is a smaller subset enriched with semantic part annotations and text labels for affordance-aware grasp generation.
  \end{minipage}
\end{table}

\section{Related Work}
\label{sec:related_works}


\subsection{Dexterous Grasp Datasets} 
Dexterous grasp datasets are crucial in the field of data-driven grasp generation. Previous works including \cite{miller2004graspit, turpin2022fastgraspd, liu2021diverse, turpin2023graspd, GRAB:2020, hasson19_obman, shreyas2020ho3d, chao2021dexycb, brahmbhatt2019contactdb, Brahmbhatt2020contactpose, YangCVPR2022OakInk, wang2024single, fan2023arctic}, provide the datasets or methods to generate datasets with diverse grasps over different objects. \cite{GRAB:2020, hasson19_obman, shreyas2020ho3d, chao2021dexycb} present datasets of grasps holding daily objects with MANO \cite{romero2017mano} hands, but the number of object types and categories is limited under 100. More recently, \cite{turpin2022fastgraspd, wang2022dexgraspnet} present large-scale datasets of grasps, with \cite{wang2022dexgraspnet} having diverse grasps on over 5000 different objects drawn from 
\cite{chang2015shapenet, savva2015shapenetsem, calli2017yale, singh2014bigbird, kappler2015leveraging, kasper2012kit, downs2022google}.
 Despite advances in creating hand grasp datasets, these works only consider single right-hand grasps. ARCTIC \cite{fan2023arctic} and H2O \cite{Kwon_2021h2o} provide dual-hand object manipulation sequence datasets; however, their grasp poses and diversity are limited because the emphasis is on motion sequences rather than grasps, and the datasets cover only a small set of objects. BimanGrasp \cite{shao2024bimanual} takes a step further, proposing a bimanual grasp synthesis algorithm to synthesize bimanual grasps. Although a large-scale dataset is collected in their work, we demonstrate that our method can generate grasps significantly larger than those of BimanGrasp, while ensuring physical plausibility.

\subsection{Data-Driven Grasp Generation}
\noindent\textbf{Semantic-agnostic grasp generation.}
Generating grasps is very challenging, given the high degree of freedom of dexterous hands. Some data-driven methods taking advantage of the large-scale datasets were proposed in previous works \cite{GRAB:2020, jiang2021graspTTA, xu2023unidexgrasp, brahmbhatt2019contactgrasp, kar2020grasp, grady2021contactopt}, synthesizing grasps based on object shapes regardless of the semantics. GraspTTA \cite{jiang2021graspTTA} is a conditional variational autoencoder-based method, which presents a two-stage framework by predicting the contacts and poses and then using the predicted contact map to refine the target grasp. UniDexGrasp \cite{xu2023unidexgrasp} generates grasp proposals by first predicting the hand root orientation, then using a normalizing flow to produce the hand translation and joint angles. The grasps are further adjusted with their refinement model. More recently, BimanGrasp \cite{shao2024bimanual} uses a DDPM to predict two-hand grasps by leveraging their synthesized data. Although these works have achieved impressive results, they only focus on the object geometry and do not consider affordances.

\noindent\textbf{Semantics-dependent grasp generation.} Whereas these methods can generate natural and realistic grasps, their neglect of human grasping intentions limits applicability in task-oriented settings. A number of semantics-driven grasp generators have emerged to address this gap, generating grasps with text intents that guide the poses \cite{wei2024grasp, dasari2023pgdm, li2024semgrasp, zhang2025dextog}.
SemGrasp \cite{li2024semgrasp} proposes a method for semantic grasp generation by aligning a discrete grasp representation with the text space, trained on a single-hand grasp-text dataset. DexTOG \cite{zhang2025dextog} introduces a language-guided diffusion-based framework for task-oriented dexterous grasp generation.
Although these works have demonstrated progress in generating semantics-dependent grasps, they suffer from a major limitation: they require all training objects to be decomposed into parts to align hand-object correspondence. Furthermore, these approaches focus exclusively on single-hand grasping and can hardly generalize to bimanual manipulation.

\begin{figure*}[!t]
  \centering
  \includegraphics[width=\linewidth]{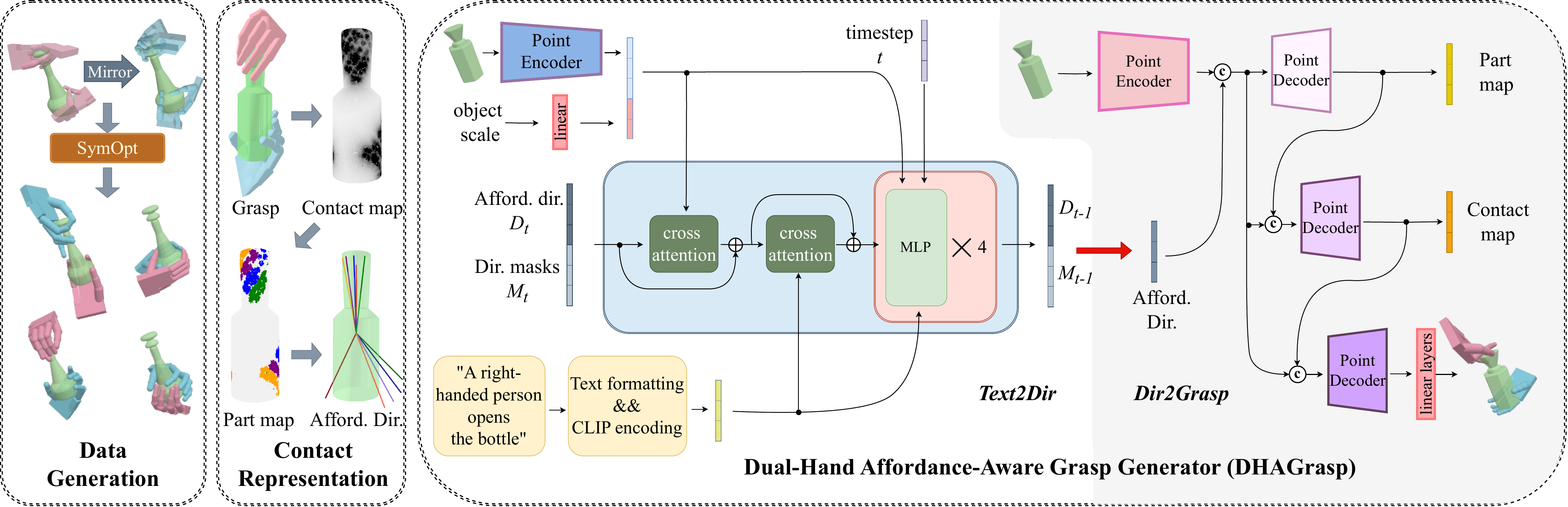}
  \caption{Our work consists of two main phases: data generation and grasp synthesis. In the data generation phase, we first mirror right-hand grasps across the pseudo-symmetry plane of the object to obtain left-hand grasp proposals. By combining the right and left grasps, we then apply an energy-based optimization scheme to construct two-hand grasps. In the grasp synthesis phase, we propose a two-hand contact representation, as illustrated in the second box, and a two-stage grasp generation pipeline, outlined in the third box. In the first stage of the generator, we design a diffusion model \textit{Text2Dir}, conditioned on the object and text, to predict the affordance directions of a grasp. The object and the predicted affordance directions are subsequently passed to our \textit{Dir2Grasp} model to generate the final grasp.
}
  \label{fig:main_fig}
\end{figure*}

\section{Dual-Hand Data Generation}
As single-hand grasping is a well-explored topic with an abundance of existing datasets \cite{wang2022dexgraspnet, chao2021dexycb, hasson19_obman, GRAB:2020}, we introduce \textit{SymOpt}, a pipeline that can convert single-hand data into a large-scale two-hand dataset. Our pipeline consists of two steps: 1) mirror the right-hand datasets to obtain left-hand grasps and combine them as initial grasp proposals, and then 2) use an energy-based optimization scheme to ensure the grasps are natural and physically plausible. In this paper, we use \cite{wang2022dexgraspnet} as the source and synthesize a two-hand grasp corpus with millions of grasps over more than 5{,}000 objects.

\subsection{Grasp Mirroring}

We first construct left-hand initial grasp proposals by mirroring the right-hand grasps from the single-hand dataset, leveraging the fact that the left hand can be regarded as a mirrored version of the right hand and observing that most objects exhibit approximate symmetry. For a given object point cloud (PCD), we first center it at the origin and then generate three mirrored variants by reflecting across the x-, y-, and z-axes, respectively. The appropriate mirror plane is determined by comparing each mirrored PCD to the original using the Chamfer Distance \cite{fan2017apoint}. Alternatively, one may employ symmetry plane detection methods such as \cite{li2023e3sym} to estimate a pseudo-symmetry plane; however, we find that simple axis flipping performs effectively in our initial experiments. Note that exact object symmetry is not required as the pipeline also produces valid grasps for asymmetric objects. 

We denote the original right-hand pose as $p_r=(\tau_r, r_r, \theta_r)$, where $\tau_r \in \mathbb{R}^3$, $r_r \in \mathbb{R}^6$, and $\theta_r \in \mathbb{R}^{22}$ represent the hand's global translation, 6D global orientation \cite{zhou2019continuity}, and local joint parameters. The right-hand PCD $V_r$ is mirrored across the selected pseudo-symmetry plane to obtain the left-hand PCD $V_l$. 
We denote the corresponding hand pose of $V_l$ at this stage as $p_l^m$. As $V_l$ is obtained through mirroring the right-hand vertices, $p_l^m$ is essentially still a right-hand pose, and we obtain the final left-hand pose $p_l$ from $p_l^m$ after the optimization step.
This allows optimization with ground-truth right-hand parameters while enforcing physical constraints on both the original and mirrored PCDs.

\subsection{Dual-Hand Grasp Optimization}
The goal of this phase is to obtain natural and physically plausible two-hand grasp poses from the initial grasp proposals. Specifically, we optimize $p_r$ and $p_l^m$ with Adam \cite{kingma2015adam} optimizer. As the right-hand grasp is already in good contact and we find the mirrored left hands usually penetrates the objects, we define the energy terms to reduce interpenetration between hand-hand and hand-object. The terms are:

\begin{equation}
\mathcal{E}_{phh}=\sum_{v\in V_l} [v\in V_r]dist(v, V_r),
\label{eq:E_phh}
\end{equation}
\begin{equation}
\mathcal{E}_{pho}=\sum_{v\in \{V_r, V_l\}} [v\in V_o]dist(v, V_o),
\label{eq:E_pho}
\end{equation}
where $[v\in V]==1$ if vertex $v$ is inside the corresponding mesh defined by $V$. The complete energy equation is:
\begin{equation}
\mathcal{E} = \lambda_{phh} \mathcal{E}_{phh} + \lambda_{pho} \mathcal{E}_{pho},
\end{equation}
where $\lambda$-s are the hyperparameters. The optimization terminates once the predefined energy thresholds are met. As the initial hand proposals are a combination of all right-left PCD pairs, many proposals suffer from severe hand-hand interpenetration and cannot be optimized. We therefore discard proposals exceeding a set interpenetration threshold. Conversely, thanks to advances in single-hand grasp data and our mirroring technique, we find that many of the proposals already meet the energy criteria prior to optimization. 

\subsection{DualHands Datasets}
With the aforementioned pipeline, we generated and selected two groups of datasets, DualHands-Full and DualHands-Sem, which can be applied for different and diverse downstream tasks.

First, we apply SymOpt to DexGraspNet and synthesize a large-scale dual-hand grasp dataset. We find that in the original DexGraspNet, there is a minor group of synthesized grasps that has hands penetrating the objects. After excluding those grasps, we obtained a dual-hand grasp dataset with over 7.8 million physically plausible grasps over 5{,}000 objects, which is more than a magnitude larger than BimanGrasp \cite{shao2024bimanual} with 150k synthesized grasps. We then select and label two-level datasets out of the synthetic data for the text-based grasp generation task. We show the dataset statistics compared to previous works in Tab. \ref{tab:statistics}.

\noindent\textbf{DualHands-Full Dataset.}
We first select a sub-dataset from the 7.8 million grasps that groups objects by their categories for the task of text-based dual-hand grasp generation.
We select the object categories based on the requirements that
each category 1) has a meaningful two-hand grasp context; 2) shares topological similarity within the same category; 3) has at least eight objects;  We end up with DualHands-Full, which consists of over 800 objects with about 1.3 million poses across nine categories, namely bottle, pistol, flashlight, hammer, headphones, lightbulb, lock, knife, and USBstick.

\noindent\textbf{DualHands-Sem Dataset.} 
As DualHands-Full is a task-agnostic grasp dataset, we construct a subset, DualHands-Sem, to enable affordance-aware grasp generation. In this subset, 157 objects are manually segmented into meaningful affordance-based parts, and text-based affordance labels are assigned to the corresponding grasps. 

For this purpose, we compute contact and part maps (detailed in the next section) to determine the contacted object part for each hand. Only grasps with more than 95\% of the hand–object contact on a single part are retained and labeled. The text label for each hand follows the format ``$\langle$chirality$\rangle$ $\langle$category$\rangle$ $\langle$part$\rangle$''. 

Because larger object parts naturally attract more grasps than smaller ones, we balance the dataset by equalizing grasp counts across affordances. From the most frequent affordance type, we retain samples closest to the average count and truncate only when it exceeds twice that of the second most frequent affordance. This balancing step prevents the generation model from being biased toward over-represented affordances or object parts.

\section{Text-Guided Grasp Generator}
Humans can identify the functional parts of previously unseen objects by recalling knowledge of familiar objects within the same category. Motivated by this ability, we introduce a text-guided grasp generator that leverages a novel dual-hand contact representation. This approach enables generalization to unseen objects using only a limited number of known objects from the same category.

\subsection{Dual-Hand Contact Representation}
For objects within the same category, segmentations are naturally related. In a typical bottle, the lid is consistently located at the narrow end, regardless of variations in overall shape. Owing to such intra-category similarities, two-hand grasps with the same affordance tend to follow consistent patterns across objects. Inspired by \cite{liu2023contactgen}, we propose a contact abstraction that can be derived from one object and readily transferred to others. This representation comprises three components: contact map, part map, and affordance directions, as illustrated in Fig.~\ref{fig:main_fig}.

\noindent \textbf{Contact map.} For a given two-hand grasp, we first compute the contact map $C_{chi}\in\mathbb{R}^{N\times 1}$ for each hand, where $chi\in[r,l]$ denotes the chirality of hands and $N$ is the number of object PCD vertices. We follow the same method as \cite{liu2023contactgen, grady2021contactopt} to compute the contact values, and each $c_{chi}^i\in[0,1]$. 

\noindent \textbf{Part map.} We compute a part map $P_{chi}\in\mathbb{R}^{N\times(B+1)}$ for each hand based on $C_{chi}$ by setting a threshold of contact. $P_{chi}$ is an array of one-hot vectors indicating which of the $B$ hand parts are in contact, and we add one dimension to designate no contact. In implementation, the hand is divided into five fingers and the palm.

\noindent \textbf{Affordance directions.} To abstract the contact affordance that can be generalized across category, we propose our affordance directions $D_{chi}\in\mathbb{R}^{B\times3}$, which are unit vectors pointing from the object PCD center to centers of each hand part contact with the object. For hand parts not in contact, we set the vector values to be zero.

\subsection{Dual-Hand Affordance-Aware Grasp Generator}

As shown in Fig.~\ref{fig:main_fig}, our dual-hand grasp generator consists of two modules, namely Text2Dir and Dir2Grasp. The Text2Dir module is a semantics-based affordance model that uses the object's shape and a text embedding to predict corresponding affordance directions. Dir2Grasp is designed to generate dual-hand grasps using the set of predicted directions as input. As obtaining the object segmentation is very time- and labour-consuming, this two-stage design allows for the learning of category-level affordances from a small number of segmented training objects, while leveraging a larger unsegmented dataset for the grasp generation phase. As our model jointly predicts the parameters of two hands, we use $F$ to denote the concatenation of $F_r$ and $F_l$, where $F_r$ and $F_l$ represent the features of the right and left hands, respectively.

\subsubsection{\textbf{Text2Dir}}
Our Text2Dir is an affordance prediction diffusion model \cite{ho2020ddpm} conditioned on the given object, text instructions, and the current timestep. We leverage the power of pre-trained PointBERT \cite{yu2021pointbert} and fix its weights as our Point Encoder at this stage. As our datasets contain object with diverse scales and pre-trained PointBERT is designed to process normalized PCDs, we also feed the logged scale of the original PCD as input to the model. 

To process the text instructions, we first use a Large Language Model (LLM) to translate any random text into the specific format $T=$``right $\langle$category$\rangle$ $\langle$part$\rangle$, left $\langle$category$\rangle$ $\langle$part$\rangle$'', then use CLIP \cite{radford2021clip} to encode $T$. In our early experiments we found that the formatting of the text strongly affects whether or not the text instruction can dictate the generation process. Directly encoding diverse texts to guide the generation can confuse the model and harm text generalizability. By utilizing LLM to format text at testing and using formatted texts at training, we simplify the text embedding for the model to learn while retaining high text generalizability.

As we set non-contact in $D_{chi}$ to be 0, we predict a binary mask $M_{chi}\in [0,1]^{B}$ along with $D_{chi}$ to indicate contactness of each hand part. At each diffusion step, we apply cross-attention \cite{vaswani2017attention} to the concatenated $D$ and $M$ with the object and text embeddings sequentially, and then pass through a sequence of Multi-Layer Perceptrons (MLPs) along with the object, text, and timestep conditions. We apply classifier-free guidance \cite{ho2022classifierfree} in our training with a condition dropping probability of 10\%. We define $D_t = \sqrt{\bar{\alpha}_t}\, D + \sqrt{1-\bar{\alpha}_t}\, \epsilon^{\text{dir}}$, $M_t = \sqrt{\bar{\alpha}_t}\, M 
    + \sqrt{1-\bar{\alpha}_t}\, \epsilon^{\text{mask}}$ and use a two-branch DDPM loss \cite{ho2020ddpm} for training, which is defined as:

\begin{equation}
\begin{aligned}
\mathcal{L}_1 &=
\mathbb{E}_{D, M, \epsilon^{dir}, \epsilon^{mask}, t} \Big[
    \lVert \epsilon^{\text{dir}} - \epsilon_\theta^{\text{dir}}(D_t, M_t, t, c) \rVert_2^2 \\
    &\quad + \lambda_{\text{mask}} \,
    \lVert \epsilon^{\text{mask}} - \epsilon_\theta^{\text{mask}}(D_t, M_t, t, c) \rVert_2^2
\;\Big],
\end{aligned}
\end{equation}

\noindent where $c$ is the conditioning object and text features.

\subsubsection{\textbf{Dir2Grasp}}
In the second phase, we design an affordance-agnostic grasp generator taking the object shape and the directions as input. As this phase does not rely on text input, the data to train the model can be obtained from the larger, unsegmented, and semantically unknown hand-object grasps. Specifically, to facilitate the learning of the spatial relationships between the vertices and contact/part maps, which are also point cloud-like features, we select the PointNet++ \cite{qi2017pointnet++} as our encoder.

The input to Dir2Grasp is the concatenation of the scaled object point cloud and vertex normals $O\in\mathbb{R}^{N\times6}$, along with the affordance directions $D$. The concatenated feature is passed through a sequence of PointNet \cite{qi2017pointnet} decoders to sequentially obtain the intermediate part map, contact map, and the final two-hand grasp parameters through linear layer heads. As our training objects have scales spanning a large range from $[0.05, 0.15]$, we observe that it can be hard for the model to regress directly on the raw global hand translations $\tau$. To solve this, we scale the global translations $\tau$ by the object diameter $d$ and predict $\tau'=\frac{\tau}{d}$ instead. The scaling maps the translations to a more evenly distributed range. 

During training, the loss of the contact map reconstruction is defined similarly to \cite{liu2023contactgen} as $\mathcal{L}_{con}=||w\odot(C-\hat{C}||)$, where $w_i=1+\lambda_wc_i$ is a weight enforcing focus on the contact region, $\odot$ denotes element-wise product, and all hats denote the reconstructed features. The part map reconstruction loss is defined as $\mathcal{L}_{part}=\mathcal{L}_{CE}(P, \hat{P})$, where $\mathcal{L}_{CE}$ is the standard cross-entropy loss. The two-hand reconstruction loss is defined as: 
\begin{equation}
\begin{aligned}
\mathcal{L}_{hand}&=||\tau'-\hat{\tau'}||+\lambda_{ori}\mathcal{L}_{geo}(r,\hat{r})\\
&+\lambda_{pose}||\theta-\hat{\theta}||_2+\lambda_V||V-\hat{V}||_2,
\end{aligned}
\end{equation}
\noindent where $\tau'\in \mathbb{R}^{3\times2}$, $r\in\mathbb{R}^{6\times2}$, $\theta\in\mathbb{R}^{22\times2}$, $V\in\mathbb{R}^{N_h\times3\times2}$ are the two hands' root translations, root 6D orientations, local pose parameters, and hand surface vertices computed from the poses, respectively. The geodemic loss is denoted as $\mathcal{L}_{geo}$. During training, we convert the 6D rotations to rotation matrices for loss computing. We finally add a penetration loss 
$\mathcal{L}_{pen}=\sum_{v\in \{V_r, V_l\}} [v\in V_o]dist(v, V_o) +\sum_{v\in V_l} [v\in V_r]dist(v, V_r)$
, where $[\cdot]$ and $dist(\cdot)$ are defined similarly to equations \ref{eq:E_phh} and \ref{eq:E_pho}, and $V_r$, $V_l$, $V_o$ are the right/left hand object surface vertices, respectively. The total loss for training is
$\mathcal{L}_2=\lambda_{con}\mathcal{L}_{con}+\lambda_{part}\mathcal{L}_{part}+\lambda_{hand}\mathcal{L}_{hand}+\lambda_{pen}\mathcal{L}_{pen}$.

\subsubsection{\textbf{Test-Time Adaptation}}
Similar to previous works \cite{jiang2021graspTTA, xu2023unidexgrasp, shao2024bimanual}, we conduct a small energy-based test-time adaptation on the generated grasps to polish their physical plausibility. The generated grasps are optimized based on the energy terms:

\begin{equation}
\mathcal{E}_{T}=\lambda_{pen}^{T}\mathcal{E}_{pen}+\lambda_{con}^{T}\mathcal{E}_{con}+\lambda_{D}^{T}\mathcal{E}_D,
\end{equation}

\noindent where $\mathcal{E}_{pen}$ is defined the same as $\mathcal{L}_{pen}$. $\mathcal{E}_{con}=AD(V_r^t, V_o)+AD(V_l^t, V_o)$, $V^t\subset V$ denotes surface vertices of the finger tips, and $AD(\cdot)$ computes the average distances between PCDs. This energy term enforces hand-object contact. The final term $\mathcal{E}_D=AD(V_r, V^I(\hat{D}_r))+AD(V_l, V^I(\hat{D}_l))$, where $V^I(\cdot)$ finds the intersection vertices on the object surface with the predicted directions $\hat{D}$, enforces semantic matching to the predicted affordance directions.

\begin{table}[!t]
  \centering
  \caption{Comparisons of the success rates validated in Isaac Gym.}
  \label{tab:success}
  \resizebox{\linewidth}{!}{%
      \begin{tabular}{lcccccc}
        \toprule
        Friction & 0.5 & 1.0 & 1.5 & 2.0 & 2.5 & 3.0 \\
        \midrule
        BimanGrasp \cite{shao2024bimanual} & 45.40 & 47.04 & 49.32 & 51.14 & 52.44 & 54.03 \\
        Ours & \textbf{95.24} & \textbf{96.17} & \textbf{97.43} & \textbf{98.29} & \textbf{99.07} & \textbf{99.76} 
        \\
        \bottomrule
      \end{tabular}
  }
\end{table}

\section{Experiments}
In this section, we conduct quantitative and qualitative experiments to evaluate the effectiveness of both our data generation algorithm SymOpt and our text-based grasp generator DHAGrasp.
Our DualHands-Sem dataset has 140 objects for training and 17 objects for testing, while DualHands-Full has 702 objects for training and 100 objects for testing.

\begin{table*}[!t]
  \centering
  \caption{Quantitative results of our method compared to previous works and our performance on different datasets.}
  \label{tab:main_table}
  \resizebox{\textwidth}{!}{%
      \begin{tabular}{l|l|l|cccccc cccccc ccc}
        \toprule
        \multirow{2}{*}{Method} & \multirow{2}{*}{Training Set} & \multirow{2}{*}{Testing Set} & \multicolumn{6}{c}{Seen} & \multicolumn{6}{c}{Unseen} & \multirow{2}{*}{$\sigma_r\uparrow$} & \multirow{2}{*}{$\sigma_l\uparrow$} & \multirow{2}{*}{$\sigma_{t/a}\uparrow$} \\
        \cmidrule(lr){4-9} \cmidrule(lr){10-15}
         & & & $Q_{1r}\uparrow$ & $Q_{1l}\uparrow$ & $Q_{1t}\uparrow$ & pen$_{ro}\downarrow$ & pen$_{lo}\downarrow$ & pen$_{rl}$vol$\downarrow$ & $Q_{1r}\uparrow$ & $Q_{1l}\uparrow$ & $Q_{1t}\uparrow$ & pen$_{ro}\downarrow$ & pen$_{lo}\downarrow$ & pen$_{rl}$vol$\downarrow$ & (cm) & (cm) & (cm) \\
        \midrule
        GraspTTA$\ast$ \cite{jiang2021graspTTA}     & DualHands-Sem & \multirow{3}{*}{DualHands-Sem} & 0.0293 & 0.0301 & / & 0.723 & 0.704 & 2.685 & 0.0287 & 0.0292 & / & 0.801 & 0.803 & 2.679 & 2.965 & 2.898 & / \\
        UniDexGrasp$\ast$ \cite{xu2023unidexgrasp}  & DualHands-Sem &                                 & 0.0404 & 0.0411 & / & \textbf{0.289} & \textbf{0.276} & 2.529 & 0.0379 & 0.0382 & / & \textbf{0.297} & \textbf{0.292} & 2.494 & 5.732 & 5.697 & / \\
        DHAGrasp (Ours)    & DualHands-Sem &                                 & \textbf{0.0412} & \textbf{0.0413} & 0.0627 & 0.427 & 0.419 & \textbf{0.392} & \textbf{0.0386} & \textbf{0.0390} & 0.0599 & 0.435 & 0.422 & \textbf{0.399} & \textbf{8.696} & \textbf{8.921} & \textbf{2.038} \\
        \midrule
        DHAGrasp (Ours)    & DualHands-Sem  & \multirow{2}{*}{DualHands-Full} & 0.0368 & 0.0371 & 0.0562 & 0.503 & 0.499 & 0.483 & 0.0374 & 0.0370 & 0.0567 & 0.493 & 0.486 & 0.477 & 8.703 & 8.674 & 2.041 \\
        DHAGrasp (Ours)  & DualHands-Full &                                 & \textbf{0.0388} & \textbf{0.0379} & \textbf{0.0583} & \textbf{0.462} & \textbf{0.458} & \textbf{0.449} & \textbf{0.0384} & \textbf{0.0391} & \textbf{0.0579} & \textbf{0.447} & \textbf{0.442} & \textbf{0.426} & \textbf{8.796} & \textbf{8.823} & \textbf{2.271} \\
        \bottomrule
      \end{tabular}
  }
  \vspace{0.4em}
  \begin{minipage}{\linewidth}
    \footnotesize
    \vspace{0.6em}
    \textit{Note}: $\ast$ are single-hand generators. We report results on both seen and unseen objects during testing. Seen and unseen in the lower part are with respect to the Dir2Grasp model on DualHands-Full, as Text2Dir for both are trained on DualHands-Sem. Best results are highlighted in bold.
  \end{minipage}
\end{table*}

\subsection{Quantitative Experiments}
\subsubsection{Data generation}
We compare our dataset generation algorithm, SymOpt, with the only previous two-hand grasp synthesis algorithm, BimanGrasp \cite{shao2024bimanual}, by the success rate of optimized grasps. A successful grasp should withhold the stability and penetration requirements simulated in a physics engine. We follow the settings of \cite{shao2024bimanual} to validate our generated grasps in Isaac Gym \cite{isaacgym}. Tab. \ref{tab:success} shows the success rate of our synthesized grasps from DualHands-Full compared with BimanGrasp under different friction coefficients. By leveraging the advances of single-hand grasp datasets, the synthesized grasps of our algorithm are almost guaranteed to be physically plausible, significantly surpassing BimanGrasp, which optimizes grasps from places far away from objects. 

\subsubsection{Grasp Generator} As our model is the first text-based two-hand grasp generator and previous works are either task-agnostic or single-handed, we first evaluate the physical plausibility of our generated results compared to previous works, then report the semantical correctness of our method. We further conduct ablation studies on the components of our pipeline to evaluate their effectiveness.

\noindent\textbf{Physical plausibility.} We evaluate the physical plausibility of our generated results with previous works. Since BimanGrasp-DDPM's implementation is unavailable, we ensure a fair comparison by comparing our pipeline against GraspTTA \cite{jiang2021graspTTA} and UniDexGrasp \cite{xu2023unidexgrasp}. We draw metrics from \cite{xu2023unidexgrasp, zhang2025dextog} and evaluate against: 1) $Q_1$, the smallest wrench needed to destabilize a grasp; 2) maximum penetration depth of a hand into the object; and 3) $\sigma$, average distances between generated hand vertices. Since GraspTTA and UniDexGrasp are single-handed grasp generators, we train two models for the right and left hands, respectively, on DualHands-Sem. We report the results in Tab. \ref{tab:main_table}. $Q_1$ and $pen.$ are computed for each of the hands. The penetration volumes of the baselines are computed by combining generated results of the right and left models. Besides the $\sigma$ of each hand, we also report a per-affordance-type average distance of both hands on our method. The results show our method achieves significantly better results on two-hand coordinance and generation diversity, while surpassing GraspTTA and achieving comparable results with UniDexGrasp on the rest of the metrics.

Besides, we also report the results of our Dir2Grasp model trained on the larger DualHands-Full dataset, which exhibits higher object geometric diversity, against the model trained on DualHands-Sem. The Text2Dir model is shared between the two and is trained on DualHands-Sem. We report the results in the lower part of Tab. \ref{tab:main_table}, which show that with only a small subset of segmented objects, our contact representation and two-stage model design enable utilizing the larger unsegmented dataset to learn the hand-object correspondence and generalize better to unseen objects.

\begin{table}[!t]
  \centering
  \caption{Results of the semantic correctness of DHAGrasp. 
  }
  \label{tab:semantics}
  \resizebox{\linewidth}{!}{%
      \begin{tabular}{l|l|ccccccc}
        \toprule
        Training set & Testing set & & dir\_s$\uparrow$ & dir\_d$\uparrow$ & grasp\_s$\uparrow$ & grasp\_d$\uparrow$ & \text{GPT-5}$\uparrow$ \\
        \midrule
        \multirow{2}{*}{DualHands-Sem} & \multirow{2}{*}{DualHands-Sem} & Seen & 96.3\% & 79.4\% & 87.2\% & 68.7\% & 73.2 \\
        & & Unseen & 95.1\% & 76.7\% & 86.8\% & 64.9\% & 70.4 \\
        \midrule
        \multirow{2}{*}{DualHands-Sem} & \multirow{2}{*}{DualHands-Full} & Seen & / & / & / & / & 65.9 \\
        & & Unseen & / & / & / & / & 66.2 \\
        \midrule
        \multirow{2}{*}{DualHands-Full} & \multirow{2}{*}{DualHands-Full} & Seen & / & / & / & / & 68.1 \\
        & & Unseen & / & / & / & / & 67.7 \\
        \bottomrule
      \end{tabular}
  }
  \vspace{0.4em}
  \begin{minipage}{\linewidth}
    \footnotesize
    \vspace{0.6em}
    \textit{Note}: The experimental setup follows the lower part of Tab. \ref{tab:main_table}.
  \end{minipage}
\end{table}

\noindent\textbf{Semantic correctness.} To evaluate the semantical correctness of our generated grasps, we report the rates of correct text-direction and text-grasp correspondence. As our grasp generation has two stages, we show the results of directions generated by Text2Dir and the final grasps, respectively. To be counted as successful text-direction correspondences, we require the generated directions to have at least two of the fingers intercepting the smaller object part, like a bottle lid, and at least have directions of the thumb and one other finger intercepting the larger part, like the bottle body. Similarly, for successful text-grasp results, we require at least two fingers touching the smaller part and the thumb and one other finger touching the large part.  Inspired by \cite{li2024semgrasp}, we further report the perception scores from 1-100 rated by LLM. For each object, we feed the rendered views of the generated grasps to GPT-5 \cite{openai2025gpt5}, asking it to rate based on the semantical correspondence, and then taking the average score. 

We report the results in Tab. \ref{tab:semantics}. To give a leveled analysis, we report the rates for at least one hand is correct, $dir_s$ and $grasp_s$, and rates for both hands, $dir_d$ and $grasp_d$. The success rates of semantics can only be calculated for DualHands-Sem and are unavailable to DualHands-Full due to the lack of object segmentations. Like Tab. \ref{tab:main_table}, the seen and unseen objects for DHAGrasp are w.r.t. Dir2Grasp training, and most objects in DualHands-Full are \textit{unseen} for Text2Dir. The results show our pipeline is able to generate affordance-aware grasps based on text instructions on unseen and unsegmented objects, achieving comparable values to segmented objects. 

\begin{table}[!t]
  \centering
  \caption{Quantitative results of ablation studies.}
  \label{tab:ablation}
  \resizebox{\linewidth}{!}{%
      \begin{tabular}{lccc ccc}
        \toprule
         & \multicolumn{3}{c}{Seen} & \multicolumn{3}{c}{Unseen}  \\
        \cmidrule(lr){2-4} \cmidrule(lr){5-7}
          & $Q_{1t}\uparrow$ &  pen$_{ho}\downarrow$ & pen$_{rl}$vol$\downarrow$ & $Q_{1t}\uparrow$ & pen$_{ho}\downarrow$ & pen$_{rl}$vol$\downarrow$ \\
        \midrule
        Ours w/o \(Inter. \& T\) & 0.0092 & 1.346 & 1.009 & 0.0073 & 1.572 & 1.138  \\
        Ours w/o \(T\) & 0.0381 & 0.729  & 0.671 & 0.0344 & 0.767 & 0.703 \\
        Ours & \textbf{0.0627} & \textbf{0.427} & \textbf{0.392} & \textbf{0.0599} & \textbf{0.435} & \textbf{0.399} \\
        \midrule
        Ours w/o \textit{Scal. Transl.} & 0.0538 & 0.446 & 0.418 & 0.0511 & 0.479 & 0.421 \\
        Ours & \textbf{0.0627} & \textbf{0.427} & \textbf{0.392} & \textbf{0.0599} & \textbf{0.435} & \textbf{0.399} \\
        \bottomrule
      \end{tabular}
    }
\end{table}

\subsubsection{Ablation Study.} We conduct ablation studies to evaluate the effectiveness of our design. In Tab. \ref{tab:ablation}, we first report the two-hand $Q_1$, max hand-object penetration depth, and two-hand penetration volumes, ablated against the test-time adaptation and whether to have intermediate guidance of contact and part maps. Specifically, we study the effect of predicting the intermediate results by Dir2Grasp compared to directly predicting the hand parameters from the object shape and affordance directions by omitting the contact and part map reconstruction losses $\mathcal{L}_{con}$ and $\mathcal{L}_{part}$. We then report the results of our pipeline trained on hand translations with and without the scaling by the object diameter, as described in section IV. The models in the experiments are all trained with DualHands-Sem.

\begin{figure}[!t]
  \centering
  \includegraphics[width=\linewidth]{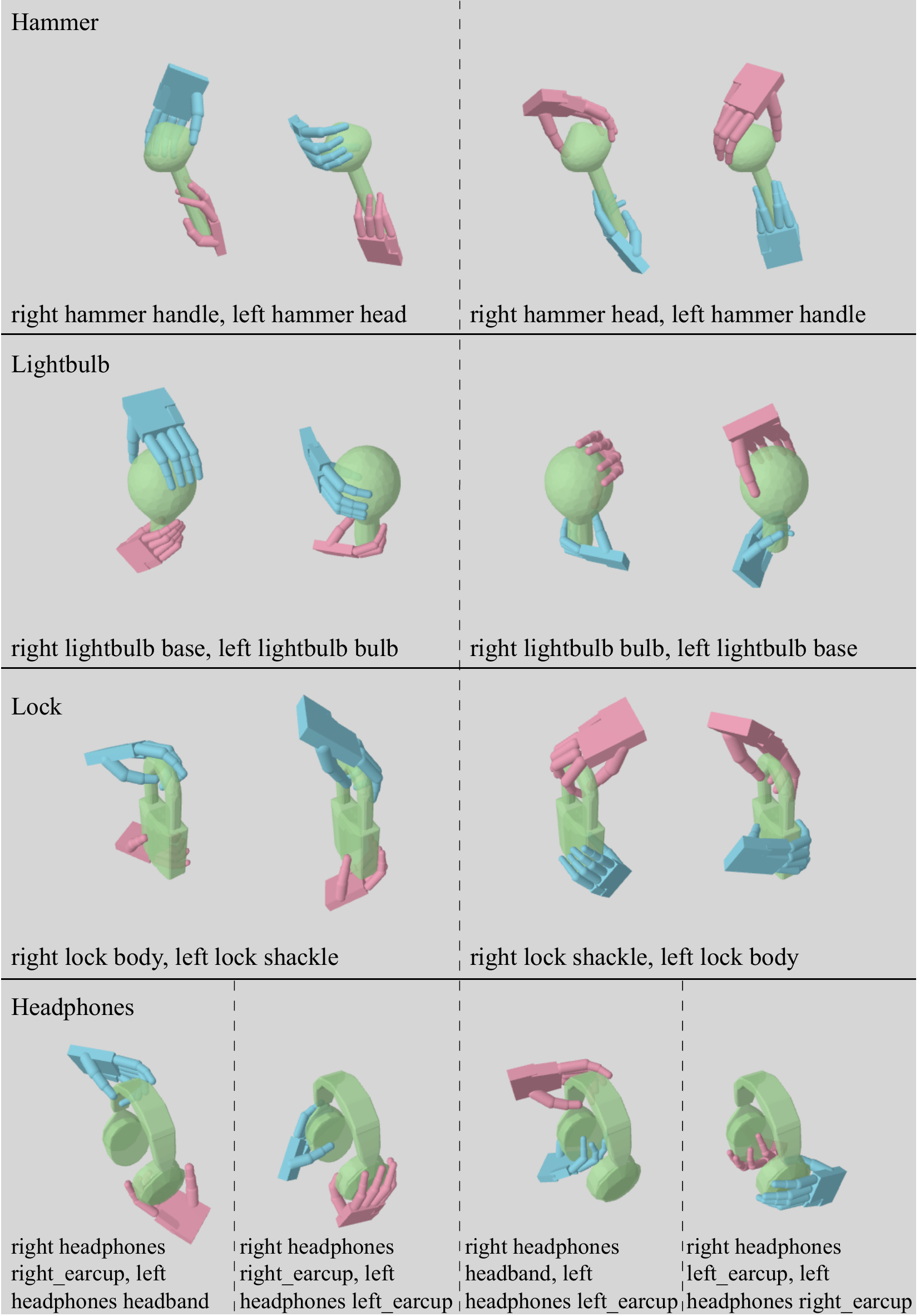}
  \caption{We present the generated results of our model on unseen objects. For the first three objects, two diverse results are shown for each affordance, while for the last object, results across four affordances are provided.}
  \label{fig:qual_1}
\end{figure}

\begin{figure}[!t]
  \centering
  \includegraphics[width=\linewidth]{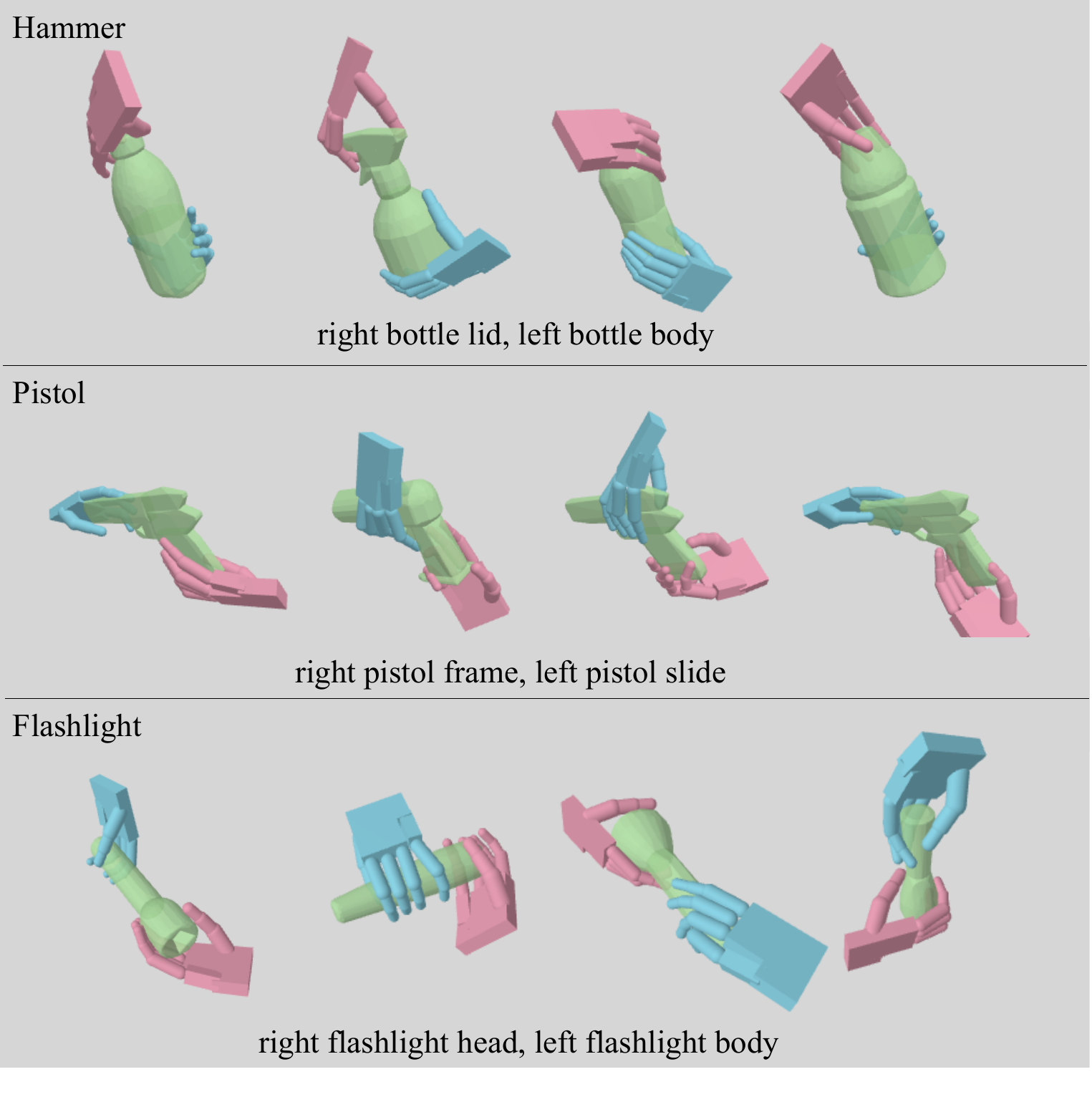}
  \caption{We show generated results of the same text instructions with different unseen objects from the same categories. The object orientations are different for the flashlight, while our model can still generate semantically meaningful results.}
  \label{fig:qual_2}
\end{figure}

\subsection{Qualitative Experiments}
In this section, we show qualitative results of our generated grasps by DHAGrasp with Text2Dir trained on DualHands-Sem and Dir2Grasp on DualHands-Full. Fig. \ref{fig:qual_1} shows the results of grasps on objects from variant categories based on different text instructions. For each text affordance, we show two diverse generation outputs for the hammer, lightbulb, and lock. We also show results of the headphones, which have more than two semantic parts, generated with four different text instructions. Fig. \ref{fig:qual_2} shows the results of our experiment with the same text affordance input but different object shapes. Note that for the flashlight results, the ground truth object orientations are different, and our pipeline can learn the correspondence between the object point clouds and their text-based affordance directions. All results demonstrated are generated with unseen objects.
\section{Conclusion}

We have introduced DualHands-Full and DualHands-Sem, the two-hand datasets generated by SymOpt pipeline, and the DHAGrasp framework for affordance-aware grasp synthesis guided by text instructions. Our SymOpt pipeline converts single-hand datasets into large-scale dual-hand data, and our two-stage DHAGrasp model leverages semantic affordances to generate grasps conditioned on object geometry and text cues. Experiments demonstrate improved grasp quality, semantic consistency, and generalization to unseen objects over prior methods.

\noindent\textbf{Limitations.} Despite these advances, our semantics primarily capture large object parts and do not yet model fine finger-level contact details. Moreover, due to the nature of our datasets, we focus on objects sized between 10–30 cm, which constrains the diversity of manipulation scenarios. In future work, we aim to extend our framework toward detailed hand–object contact control with textual guidance and to enable dual-hand cooperation on larger objects. These directions represent important steps toward more versatile and semantically meaningful robotic manipulation.

{
    \small
    \bibliographystyle{IEEEtran}
    \bibliography{main}
}

\end{document}